\documentclass[conference]{IEEEtran}
\IEEEoverridecommandlockouts
\usepackage{cite}
\usepackage{amsmath,amssymb,amsfonts,amsopn}
\usepackage{algorithmic}
\usepackage{algorithm}
\usepackage{graphicx}
\graphicspath{{figures/}}
\usepackage{textcomp}
\usepackage{xcolor}
\def\BibTeX{{\rm B\kern-.05em{\sc i\kern-.025em b}\kern-.08em
    T\kern-.1667em\lower.7ex\hbox{E}\kern-.125emX}}
\usepackage{cleveref}
\usepackage{float}
\usepackage{subcaption}
\usepackage{pgffor}
\usepackage{epstopdf}
\usepackage{tikz}
\usepackage{venndiagram}
\usepackage{algorithmic}
\usepackage{comment}
\ifpdf
  \DeclareGraphicsExtensions{.eps,.pdf,.png,.jpg}
\else
  \DeclareGraphicsExtensions{.eps}
\fi

\begin{document}

\title{Feature Reduction Method Comparison Towards Explainability and Efficiency in Cybersecurity Intrusion Detection Systems
\thanks{Research in large conducted thanks to funding from National Security Agency grant H98230-22-1-0017 and National Science Foundation award 1719498.
}
}

\author{\IEEEauthorblockN{Adam Lehavi}
\IEEEauthorblockA{\textit{Viterbi School of Engineering} \\
\textit{University of Southern California}\\
Los Angeles, CA 90089 \\
alehavi@usc.edu}
\and
\IEEEauthorblockN{Seongtae Kim}
\IEEEauthorblockA{\textit{Department of Mathematics and Statistics} \\
\textit{North Carolina Agricultural and Technical State University}\\
Greensboro, NC 27411 \\
skim@ncat.edu}
}

\maketitle

\begin{abstract}
In the realm of cybersecurity, intrusion detection systems (IDS) detect and prevent attacks based on collected computer and network data. In recent research, IDS models have been constructed using machine learning (ML) and deep learning (DL) methods such as Random Forest (RF) and deep neural networks (DNN). Feature selection (FS) can be used to construct faster, more interpretable, and more accurate models. We look at three different FS techniques; RF information gain (RF-IG), correlation feature selection using the Bat Algorithm (CFS-BA), and CFS using the Aquila Optimizer (CFS-AO). Our results show CFS-BA to be the most efficient of the FS methods, building in 55\% of the time of the best RF-IG model while achieving 99.99\% of its accuracy. This reinforces prior contributions attesting to CFS-BA's accuracy while building upon the relationship between subset size, CFS score, and RF-IG score in final results.
\end{abstract}

\section{Introduction}
\label{sec:introduction}

Cybersecurity is a growing field with increasing necessity and prominence. In 2022 thus far, there have been 1.364 billion identified malware programs\cite{noauthor_malware_nodate}.
The average cost of a data breach is \$4.24 million, with an 80\% cost difference where security AI was deployed through either Machine Learning (ML) or Deep Learning (DL)\cite{noauthor_cost_2021}. Threat detection, or finding malicious activity, is one of the largest components of the cybersecurity field. An intrusion detection system (IDS) is a model that detects these threats through various means \cite{xin_machine_2018}. A network-based IDS (NIDS) monitors network connections to look for malicious traffic \cite{sarker_cybersecurity_2020}. As networks are typically the source of more damaging and costly problems, spanning company and organization data, building an NIDS has importance in current research, and will be the focus of this paper.

Intrusion detection systems are typically divided into two separate classes based upon their approach and the type of attacks they aim to cover. Signature-based IDS relies on known attacks; it generates patterns from past or given data and then uses those set patterns to sift through current data, such as an antivirus package\cite{sarker_cybersecurity_2020}. Anomaly-based IDS aims to find dynamic patterns to group data and search for deviations \cite{sarker_cybersecurity_2020}. Hybrid approaches combine both methods. While signature-based systems have guarantees on detecting common malicious activity, they perform poorly on attacks that deviate from those patterns. In addition, they often take large amounts of labeled data beforehand, and need constant updating. As such, anomaly-based detection can achieve better results for large amounts of data with unknown correlations.

NIDS are fundamentally built upon data from the network. In research, data to test particular models and methods comes from datasets of real or simulated network data. The most common datasets include NSL-KDD, KDD-Cup'99, UNSW-NB15, CIC-IDS2017, Kyoto, and CSE-CIC-IDS2018 \cite{farhan_performance_2020}. Of these sets, CSE-CIC-IDS2018 is the most current, meaning it is the most applicable. We therefore selected this dataset to be able to add a point of comparison with needed metrics \cite{zhou_building_2020}. The dataset is useful for practice, because it contains a variety of attacks including zero-day attacks, or ones that typically happen when a network is initially set up and open to user activity \cite{sarker_cybersecurity_2020}. Since it is not as common as CIC-IDS2017 and lacks many results, it is a good target dataset\cite{farhan_performance_2020}.

In building these intrusion detection systems, we will look to use ML and DL methodology. ML concerns itself with statistical methods that construct or evaluate patterns from known functions and behavior \cite{james_introduction_2021}. Within all possible ML methods, we need to look towards classification methods that specify if a user is attacking and how, as well as one that can work regardless of our class distribution, as we fundamentally need to explore attacks of all types, and CSE-CIC-IDS2018 contains very skewed data. This leads us to preferring K-nearest neighbors and Random Forest (RF) \cite{james_introduction_2021}. K-nearest neighbors is a non-parametric learning algorithm classifying new points to the majority class of the \(K\) closest points. Because K-nearest neighbors is difficult to extract information from, and the primary motivation of this research is to compare methodologies based on a variety of factors including relative importance of predictors, we will not use it. We will use RF, however, for a few reasons. For data distributed with many extreme points, using a method like RF that does not rely upon weights for given inputs but rather specific boundaries may give it an edge in performance. Because RF has a relatively good balance of low variance to low bias, as well as having much better results than a standard decision tree, all while still being relatively easily interpretable, we will use it \cite{james_introduction_2021}.

Deep Neural Networks (DNN) are an offshoot of DL with the goal of modeling complex nonlinear relationships using chains of neurons and activation functions. With regards to our work, they are a strong tool to choose and incorporate because of their fast training time given the scope of our data and consistently strong performance when compared to nearly all ML methodology \cite{leevy_survey_2020}.

In exploring our results, we will display the information in a manner that is as transparent and digestible as possible. Similar to many papers, we include the accuracy, precision, F1 score, and false alarm rate (FAR). Most  other metrics can be determined as a derivative of these, and they provide a foundation by which to compare results \cite{leevy_survey_2020, mahdavifar_application_2019}. While \cite{farhan_performance_2020} achieves results for our dataset using DNN architecture, we achieve even better results and thus choose to include it in conjunction with RF.

Feature selection, which aims to feed a model a subset of the original features to improve, is the central focus of the paper. These improvements can include time to build a model as well as performance results such as accuracy, precision, F1 score, and FAR. There are three types of feature ranking methods commonly used. Filters use predefined criteria or patterns in the data, wrappers build many models to compare usefulness of given features, and embedded methods rely upon training a model that then defines features for another model \cite{akashdeep_feature_2017}.
For the sake of being able to compare more modern optimizer filter methods, we will use both other filter methods and embedded methods as benchmarks, since wrapper methods both have better accuracy and much larger build times.
A standard method for embedded feature selection is RF information gain (IG). Because of its large amount of time to compute, it is useful but limited in applicability \cite{li_building_2020}. The filter methods to be compared are the correlation based feature selection Bat Algorithm (CFS-BA), and Aquila Optimizer (CFS-AO).

Correlation based feature selection is built upon the idea that an outperforming subset of features can be determined purely through correlation between variables. The score derived is one that aims to maximize correlation between predictors and final classes while minimizing correlation between pairs of predictors \cite{zhou_building_2020}. This methodology performs much faster than any embedded method, and can give more intuitive reasoning into the decision making process, leading us to choose it. The only issue is that we cannot explore every possible subset, and as such need to use an optimizer method to find the closest subset to the global maxima score.

Bat Algorithm (BA) was proposed by Yang in 2010 as a metaheuristic optimization algorithm to explore a space mimicking the behavior of microbats using echolocation \cite{yang_bat_2012}. Most metaheuristic algorithms have two phases; exploration attempts to discover the widest range of the feature space, and exploitation aims to discover the best local solution within a given region. Unlike most modern metaheuristic algorithms, BA gradually switches between exploration and exploitation using a self-tuning hyperparameter \cite{yang_bat_2013}. CFS-BA has performed well for CIC-IDS2017, but does not have CSE-CIC-IDS2018 results and lacks results for DNN and RF models \cite{zhou_building_2020}.

The Aquila Optimizer was introduced as a faster and potentially more efficient metaheuristic algorithm than prior methodology. While slower to converge onto a maxima, it has been shown to outperform BA when used in feature selection on cybersecurity benchmark datasets \cite{fatani_advanced_2022}. Because of its promising nature, the algorithm will be used and compared to BA with CFS scoring. AO has performed well for CIC-IDS2017, but used a different fitness than CFS and does not have CSE-CIC-IDS2018 results \cite{fatani_advanced_2022}.

The paper is organized as follows: our motivation for the particular fields, tools, and algorithms used are in \cref{sec:introduction},
particular mathematical and technical details regarding the implementation of our research are in
\cref{sec:methodology},
information specific to our dataset and its distribution and processing are in \cref{sec:data},
experimental results are in \cref{sec:results},
interpretation of results and limitations of the study are in \cref{sec:discussion},
and a summary of the work's main achievements as well as future work are in \cref{sec:conclusion}.

\section{Methodology}
\label{sec:methodology}
Our methodology will discuss the specifics of how we will build and tune RF and DNN models for feature selection comparison.
\subsection{Classification Algorithms}
\label{subsec:classification}
Classification algorithms work by feeding a set of inputs through a model to output a final, categorical variable. The methods we will explore for this are RF and DNN.

\subsubsection{Random Forest}
\label{subsubsec:meth_clas_rf}

\begin{figure}[htbp]
    \centering
    \includegraphics[width=0.4\textwidth]{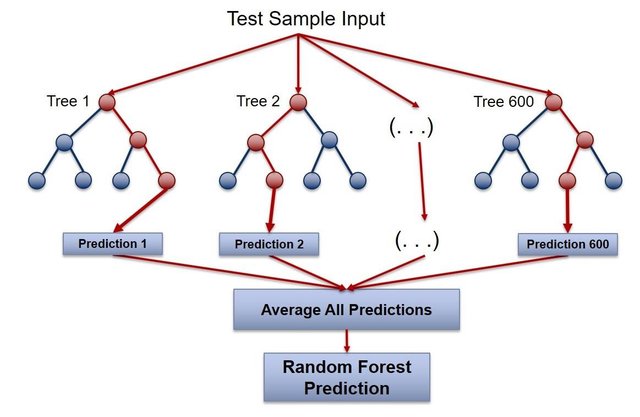}
    \caption{Visualizations of an example RF courtesy \cite{blakely_evaluation_2018}.}
    \label{fig:decision_tree_and_rf_diagrams}
\end{figure}

The basic building block for RF is the decision tree. Visualization is in \cref{fig:decision_tree_and_rf_diagrams}. The decision tree is a model that classifies data points by putting them through a series of binary decision boundaries and then assigning them to the same class as the majority of the training points within a final bucket. The equation used to choose decision splits in this paper, the Gini Index \cite{james_introduction_2021}, is defined as \cref{eq:gini}. Entropy can also be used, defined as \cref{eq:entropy} \cite{james_introduction_2021}.

\begin{align}
    G &= \sum_{k=1}^K\hat{p}_{mk}(1 - \hat{p}_{mk}) \label{eq:gini} \\
    D &= -\sum_{k=1}^K\hat{p}_{mk}\log\hat{p}_{mk} \label{eq:entropy}
\end{align}

Both indexes are modeled to be a measure of total variance across all classes, with a smaller value indicating a better decision tree. Here, \(K\) represents the number of classes, and \(\hat{p}_{mk}\) represents the proportion of training observations in the \(m\)th region from the \(k\)th class. The indexes takes a smaller value for \(\hat{p}_{mk}\) closer to zero or one, meaning it acts as a measure of \emph{purity} for a given node.

Splits are made to minimize \cref{eq:gini}. For a standard decision tree, every split can compare all possible predictors; however, when used in conjunction with bootstrapping, this increases variance and decreases accuracy \cite{james_introduction_2021}. To work around this, RF forces each split to only compare one of $m = \sqrt{p}$ out of the total \(p\) predictors.

In Random Forest, all inputs are used to train a set of bootstrapped decision trees. This means that some number of trees, set in this case to be 100, are constructed from sets with the same amount of points as the original data set, made by randomly selecting points from the base set with replacement. Each of these decision trees will have decision splits added but never removed, and reach a depth $d$.  We tested the $d$ range of [2,5,10,20,40,100,200] building RF models for each and calculating the OOB score, and found $d = 20$ ideal for our performance. Out-of-bag works by plugging the training set back into the finalized model and recording the fraction of the training set that ends up correctly labeled. This simulates predicted accuracy to a certain degree, and thus is a substantial enough metric for our use.

Furthermore, we can use \cref{eq:entropy} to determine the information gained at each split by each variable. Working with the average value of all of these splits over each predictor gives us our information gain (IG) scores, which we can use to determine feature subsets.

\subsubsection{Deep Neural Networks}
\label{subsubsec:meth_clas_dnn}

Neural Networks, and DNN more specifically, consist of an input layer \(X_1,\ldots,X_n\), describing all predictors in numerical terms, multiple hidden layers, consisting of fully connected nodes passed to the next layer through nonlinear activation functions, and an output layer. In the case of DNN for the sake of classifying, the process is as follows:

A hidden layer $n$ with $D$ nodes with prior layer $m$ having $C$ nodes can be represented as
\[A_d^{(n)}=h_d^{(n)}(X)=g(w_{d_0}^{(n)}+\sum\nolimits_{j=1}^pw_{d_j}^{(n)}A_c^{(m)})\]
for some activation value \(A_d^{(n)}\) for node $d$ defined through activation function $h$ expanded for some nonlinear function \(g\) using weights \(w\) for all \(p\) predictors. For the first layer, $A_c^{(m)}$ can be substituted with $X_j$.
This is then repeated until layer $n'$ with $D'$ nodes and piped to \(K\) different linear models
\[Z_k = \beta_{k_0}+\sum\nolimits_{d'=1}^{D'}\beta_{k_{d'}}h_{d'}^{(n')}(X)=\beta_{k_0}+\sum\nolimits_{d'=1}^{D'}\beta_{k_{d'}}A_{d'}^{(n')}.\]
This value is then taken and put through the \emph{softmax} activation function \cref{eq:softmax} for multiclass classification and the \emph{sigmoid} function \cref{eq:sigmoid} for binary classification \cite{james_introduction_2021}. Afterward it is classified to the \(f_k(X)\) with the highest probability.
\begin{align}
    f_k(X)=Pr(Y=k|X) &= \frac{e^{Zk}}{\sum\nolimits_{l=0}^Ke^{Zl}} \label{eq:softmax} \\
    f_k(X)=Pr(Y=k|X) &= \frac{e^Z}{1+e}=\frac{1}{1+e^{-Z}} \label{eq:sigmoid}
\end{align}

All other hidden layer activation functions are chosen to be ReLU (rectified linear unit), defined as $A(z) = (z)_+$ that returns $0$ for negative inputs and $z$ otherwise, as it achieves far better performance than most other alternatives \cite{james_introduction_2021, noauthor_deep-learning_nodate}.

\begin{figure}[H]
    \centering
    \includegraphics[width = 0.4\textwidth]{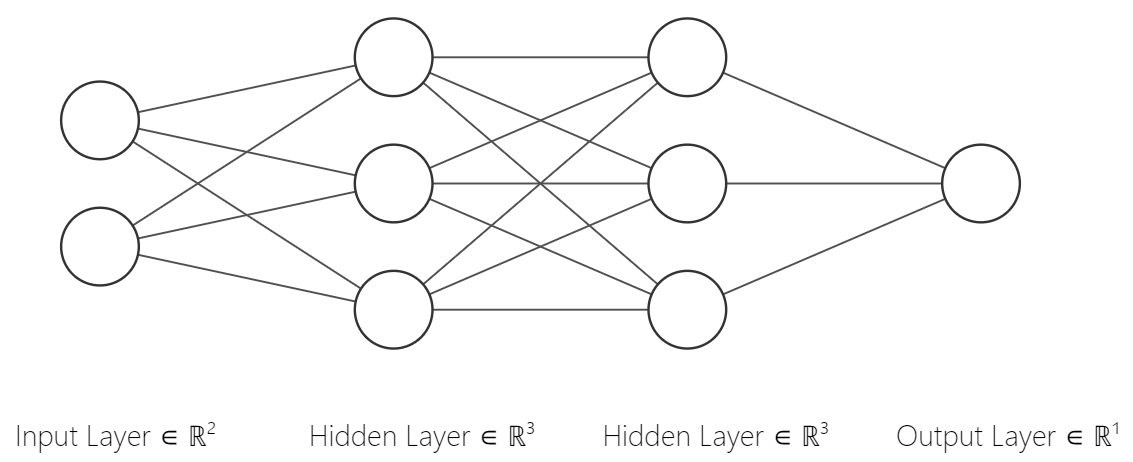}
    \caption{Deep Neural Network Visualization for 2 inputs and 1 output}
    \label{fig:dnn_visualization}
\end{figure}

The number of neurons and hidden layers, while recommended to be large, may perform best when minimized given the sparsity of smaller classes \cite{noauthor_deep-learning_nodate}. As such, we compared two methodologies. The first assumes that we have enough information to warrant a complex model. This model contains three hidden layers, all of size 100. We will refer to it as the 100-100-100 model to reinforce this. The second model will be built on the assumption that we have a lack of information for the size of our data. This encourages a structure that funnels information down to the final layer, and so we will use a hidden layer of size 50 first and then of size 25. This will be the 50-25 model. After these hidden layer passes for each respective model, the classification will be done.

This training is done in batches, as the results and weights are updated for each run. Common batch sizes are 32, 64, 128, and 256. The batch size typically makes minimal difference and is limited by a resource's RAM, and so we choose to use batch size 32 for the sake of maximizing result reproducability.

Overall, the 100-100-100 model underperformed the 50-25 model in build time by large margins and accuracy by notable margins for all of the categorical models. This led us to compare RF and DNN using the 50-25 structure.

\subsection{Correlation Feature Selection}
\label{subsec:meth_cfs}
CFS is a filter method that analyzes the fitness of a given subset of features based solely on the intercorrelation between features. This is done through Spearman's correlation, which works for numerical values calculating colinearity. To make the formula work for our classification, we used one-hot encoding to get individualized scores.

The score for a given subset, \(s\), can be calculated as \cref{eq:cor} \cite{hall1999correlation}.
\begin{equation}\label{eq:cor}
    M_s=\frac{k\bar{r_{cf}}}{\sqrt{k+k(k-1)\bar{r_{ff}}}}
\end{equation}
\(M_s\) is the score, valuing a higher number as a better subset, \(k\) is the number of predictors for a given subset, \(\bar{r_{cf}}\) is the average absolute intercorrelation between predictors and final features, and \(\bar{r_{ff}}\) is the average absolute intercorrelation between predictors. We look at the average \emph{absolute} intercorrelation, so that the formula rewards inverse and direct relations the same.

\subsection{Bat Algorithm}
\label{subsec:meth_ba}
Our main goal is to use a metaheuristic optimizer to explore a feature space for a given best point. In our case, the main metaheuristic optimizer is BA, the feature space is all possible subsets of the 70 preprocessed predictors, and the best point is the one with the maximum value from \cref{eq:cor}.

 We define some number of bats and instantiate each one according to \cref{eq:new_bat}. Uniform distribution between values \(a\) and \(b\) is denoted as \(U(a,b)\). \(x_i^0\) represents the initial position for bat \(i\) out of a total of \(n = 100\) bats, with an \(x\) value of \([0,0.5]\) at position \(i\) meaning a rejection of the predictor in location \(i\) and a value of \([0.5,1]\) representing acceptance. \(v_i^0\) represents the initial velocity, \(f_i^0\) represents the initial frequency, \(A_i^0\) represents the initial loudness, and \(r_i^0\) represents some initial random value.
\begin{equation}\label{eq:new_bat}
    \begin{aligned}
        x_i^0 &= \begin{bmatrix}x_1\\ x_2\\ \cdots\\ x_k \end{bmatrix} = 
        \begin{bmatrix}U(0,1)\\ U(0,1)\\ \cdots\\ U(0,1) \end{bmatrix}\\
        v_i^0 &= \begin{bmatrix}v_1\\ v_2\\ \cdots\\ v_k \end{bmatrix} = 
        \begin{bmatrix}U(-1,1)\\ U(-1,1)\\ \cdots\\ U(-1,1) \end{bmatrix}\\
        f_i^0 &= U(0,0.1),\:\: A_i^0 = U(1,2),\:\: r_i^0 = U(0,1)
    \end{aligned}
\end{equation}

We then define the best point \(x_b\) to be the position that correlates to the highest value for \cref{eq:cor}. Then for each epoch \(t\), with \(t_{max} = 1000\), we first update \(v_i^t\) according to \cref{eq:up_f}, \cref{eq:up_v}, and \cref{eq:clamp_v}.
\begin{align}
        f_i^t &= U(0,0.1)\label{eq:up_f}\\
        v_i^t &= v_i^{t-1} + (x_i^{t-1} - x_b)f_i^t \label{eq:up_v}\\
        v_i^t &= \max\left(\min\left (v_i^t,1 \right ), -1\right ). \label{eq:clamp_v}
\end{align}

Then we update \(x_i^t\) according to \cref{eq:x_up1} or \cref{eq:x_up2}. \(A_i^t\) represents the loudness.
\begin{align}
    x_i^t &= \max\left(\min\left (x_i^{t-1} + v_i^t, 1\right ), -1\right )\label{eq:x_up1}\\
    x_i^t &= \max\left(\min\left (x_b + \begin{bmatrix}
        U(-0.01,0.01)\\ U(-0.01,0.01)\\ \cdots \\ U(-0.01,0.01) \end{bmatrix}A_i^t, 1\right ), -1\right ) \label{eq:x_up2}
\end{align}
After this, we update our frequency and loudness using \cref{eq:up_A} and \cref{eq:up_r} where \(\alpha\) and \(\gamma\) are constants, both of which can be set to 0.95.
\begin{align}
    A_i^{t+1} &= \alpha A_i^t \label{eq:up_A}\\
    r_i^{t+1} &= r_i^0(1-e^{-\gamma^t}\label{eq:up_r})
\end{align}

After determining all of these values, we can run the Bat Algorithm. The main idea of this algorithm is to transition from exploration to exploitation gradually through self-tuning behavior. We can now create \cref{alg:bat}, based largely on \cite{yang_bat_2012, zhou_building_2020, yang_bat_2013}.
\begin{algorithm}[htbp]
\caption{Bat Algorithm}
\label{alg:bat}
\textbf{Input}: \quad Correlation matrix produced from training dataset\\
\textbf{Output}: \quad Selected best feature subset \(x_b\)
\begin{algorithmic}[1]
    \STATE{Initialize $n$ bats using \cref{eq:new_bat}}
    \STATE{$x_b \gets 0, M_b \gets 0$}
    \FOR{$i \in [1,n]$}
        \STATE{Calculate $M_i$ for subset $s_i$ determined by bat $x_i$ using \cref{eq:cor}}
        \IF{$M_i > M_b$}
            \STATE{$x_b \gets x_i, M_b \gets M_i$}
        \ENDIF
    \ENDFOR
    \STATE{$t \gets 1$}
    \WHILE{$t \leq t_{max}$}
        \FOR{$i \in [1,n]$}
            \STATE{Update $f_i$ using \cref{eq:up_f}}
            \STATE{Update $v_i$ using \cref{eq:up_v} and then \cref{eq:clamp_v}}
            \IF{$r_i \geq U(0,1)$}
                \STATE{Update $x_i$ using \cref{eq:x_up1}}
            \ELSE
                \STATE{Update $x_i$ using \cref{eq:x_up2}}
            \ENDIF
            \STATE{Calculate $M_i$ using \cref{eq:cor}}
            \IF{$A_i > U(0,1)$ \AND $M_i > M_b$}
                \STATE{$x_b \gets x_i, M_b \gets M_i$}
                \STATE{Update $r_i$ using \cref{eq:up_r}}
                \STATE{Update $A_i$ using \cref{eq:up_A}}
            \ENDIF
        \ENDFOR
        \STATE{$t \gets t + 1$}
    \ENDWHILE
    \RETURN $x_b$
\end{algorithmic}
\end{algorithm}

\subsection{Assessment Metrics}
\label{subsec:assessment}
To determine the fitness and results we need to clarify the metrics our models will use.

\subsubsection{Binary Classification}
\label{subsubsec:binary}
When we get our results for both RF and DNN, we get them in the form of a confusion matrix. This matrix describes the amount or ratio of what the data's class actually is compared to what it is described as. In the case of binary classification describing if a data point is benign or malicious, there are universal metrics. We will be describing an attack as positive and a benign point as negative. As taken from \cite{xin_machine_2018}, we will denote True Positive (TP), False Negative (FN), False Positive (FP), and True Negative (TN). Standard used metrics are defined below as
\begin{align}
    Accuracy &= \frac{TP+TN}{TP+TN+FP+FN} \label{eq:accuracy}\\
    Precision &= \frac{TP}{TP+FP}\label{eq:precision}\\
    Recall &= \frac{TP}{TP+FN}\label{eq:recall}\\
    FAR &= \frac{FP}{FP+TN}\label{eq:far}\\
    F1 &= 2\left ({Recall}^{-1} + {Precision}^{-1} \right )^{-1}\label{eq:f1}
\end{align}
with FAR denoting the False Alarm Rate.

The choice of these metrics is mainly a result of suggested and common metrics from \cite{ahmad_network_2021} and \cite{leevy_survey_2020}.

\subsubsection{Multiclass}
\label{subsubsec:multiclass}

For multiclass classification, we want to use the same metrics, but there is no longer a single target class. As such, all metrics need to treat every class as a target and average in some way. Accuracy stays consistent as the ratio of correctly labeled points to all points. For the other metrics, we consider using either micro, macro, or weighted values.

Micro averages compute the values for the entire table, being blind to the individual classes. Because of this, the micro equations for F1 and Precision are equal and equate to Accuracy, defined as \cref{eq:macro_accuracy}.

\begin{equation}\label{eq:macro_accuracy}
    Accuracy = \frac{Correctly\:Labeled\:Points}{All\:Points}
\end{equation}

Macro averages compute the values on a class by class basis and then average all the independent values.
Weighted averages work similar to macro averages, but instead weigh the score from each class by the proportion of that class to the total. This means that weighted averages end up giving us slightly less information as they mimic overall accuracy. Furthermore, our goal with using CFS-BA is to focus on each class with equal emphasis since we use one-hot encoding. Therefore, macro scores will tend to give us results more reflective of how we performed compared to our expectations.

\section{Data Description}
\label{sec:data}

\begin{figure}[htbp]
  \centering
  \includegraphics[width = 0.5\textwidth]{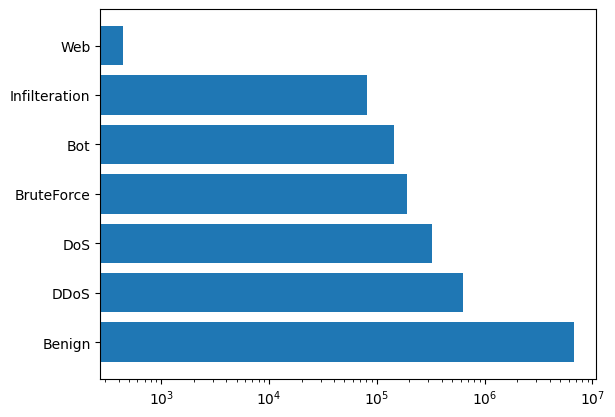}
  \caption{Log Scaled Frequency of Classes}
  \label{fig:log_cat_dist}
\end{figure}

The dataset used was created by the Communications Security Establishment (CSE) and the Canadian Institute for Cybersecurity (CIC) in 2018 to simulate network data for the sake of aiding in the creation and testing of NIDS construction \cite{noauthor_ids_nodate, noauthor_realistic_nodate}. It contains simulated data for various attacks over 10 days, with each day having a different distribution. The full distribution on a log scale is described in \cref{fig:log_cat_dist}. Although there are 6 types of attacks, they are labeled as and can be expanded to 13 types of unique attack signatures. There are 83 numerical inputs, both continuous and discrete inputs, and each day has at least 79 inputs.

\subsection{Data Preprocessing}
\label{subsec:data_pre}
To get the data into a more suitable format, it underwent preprocessing. The first thing done to the data was to remove \{\textbf{Dst.IP, Flow.ID, Src.IP, Src.Port}\}, features that are present for one of the days but not others. Our model will be constructed for all of the days, and thus the large amount of NA values that the inclusion of these would generate would cause issues. 

Of the remaining 79 predictors, 1 had NA values and 2 had $\infty$ values. There are only 59,721 rows with NA and 36,039 rows with $\infty$ out of 16.17 million.
As such, it is better to remove the points instead of the predictors.
Additionally, \{\textbf{Timestamp}\} can be removed. It is unique to the simulated environment and the inclusion of it would lead to a model built for past data, and not projecting towards future data. While it may seem like \{\textbf{Dst.Port}\} is similar to this, a model can be based on banning or including certain connections, so it is included.
This leaves us with 78 predictors, of which 8 more can be removed as they have uniform values for all days.

Data is then normalized using MinMax normalization in the range of [0,1], as data is not normally distributed enough for Gaussian normalization, and is typically skewed about an extreme or completely non-negative, not befitting [-1,1] MinMax normalization. The full distribution for each variable can be found through \cite{kaggle_ids_nodate}. Most of the predictors have highly skewed and nonlinear data, with a spike toward some extreme.

For our model analysis, we used a test-train split of 50/50 because our model is large enough such that a smaller portion of it can be used to train and the weaker classes are small enough such that we need both test and train to be a significant portion of the data to get accurate results.

\section{Results}
\label{sec:results}

Time results were achieved through Python ran in Google Colab Pro+. CFS-BA, CFS-AO, and RF-IG were generated using both \cite{scikit-learn} and \cite{thieu_nguyen_2020_3711949}. The RF results were generated using scikit-learn \cite{scikit-learn}. The DNN results were generated using TensorFlow \cite{tensorflow2015-whitepaper}.

The computer by which the hyperparameter results were achieved and compared in terms of run time has a Nvidia GTX 1650 GPU, 32 GB of RAM, and has a Intel Core i7 processor. Finalized building and running of models was done through Google Colab Pro +, which utilized up to 53 GB of RAM for the run sessions. 

\subsection{Correlation}
\label{subsec:cor_res}

\begin{figure}[htbp]
  \centering
  \includegraphics[width=0.4\textwidth]{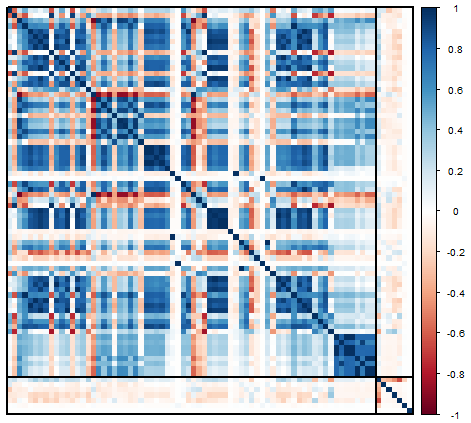}
  \caption{Correlation Heatmap for Full Feature Set}
  \label{fig:initial_corr}
\end{figure}

Correlation structure is calculated according to the methodology in \cref{subsec:meth_cfs}. After calculating our correlation, we stored it in a matrix and then added a dividing line indicating where the features end and the classes begin. The matrix is heat-encoded and shown in \cref{fig:initial_corr}. The black lines crossing the figure represent the divide between predictors and classes, and the diagonal line through the figure represents each variable's intercorrelation with itself, which is always 1. The main issue is that most of the regions that have strong feature to class intercorrelation also have strong feature intercorrelation. To maximize the result of $M_s$ from \cref{eq:cor}, we want to find features that have the lightest average coloration beneath the diagonal line in the top left rectangle and the darkest coloration in the bottom left rectangle. Specific predictor names are omitted for the sake of readability and interpretability. The original $M_s$ score is \textbf{0.119} for multiclass and \textbf{0.248} for binary. The original information gain sum (IG) score is 1 for both sets by definition.

\begin{figure}
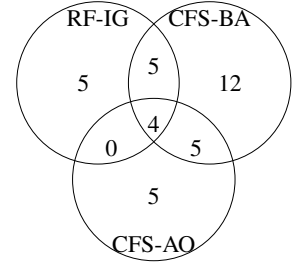

    \centering
    \scalebox{0.9}{
    \begin{venndiagram3sets}[labelOnlyA=5, labelOnlyB=12, labelOnlyC=5,labelOnlyAB=5,labelOnlyAC=0,labelOnlyBC=5,labelABC=4,labelA=RF-IG,labelB=CFS-BA,labelC=CFS-AO,showframe=false]
    \end{venndiagram3sets}
    }
    \caption{Venn Diagram of Categorical Feature Subset Overlap}
    \label{fig:venn}
\end{figure}

After running CFS-BA, CFS-AO, and RF-IG for our tuned hyperparameters, we are left with features that give us $M_s$ scores and IG scores shown in \cref{tab:reduced_subset_scores} and \cref{fig:venn}. The four features common to all  methods are \{\textbf{Fwd.Pkt.Len.Mean, Flow.Pkts., Init.Fwd.Win.Byts, Fwd.Seg.Size.Min}\}. Note that for \cref{tab:reduced_subset_scores}, \cref{tab:dnn_50_perf_res}, and \cref{tab:rf_perf_res}, cat. denotes categorical, bi. denotes binary, and time denotes build time.

\begin{table}[htbp]
\footnotesize
\caption{Reduced Feature Subset Scores}\label{tab:reduced_subset_scores}
\centering
\resizebox{0.8\columnwidth}{!}{
  \begin{tabular}{|l|c|c|c|} \hline
    Methodology&K&CFS & IG \\ \hline
    Cat. CFS-BA &26&0.147 & 0.630 \\ \hline
    Cat. CFS-AO & 14 & \textbf{0.159} &0.350 \\ \hline
    Cat. RF-IG&14&0.136& \textbf{0.678} \\ \hline \hline
    Bi. CFS-BA & 8 & 0.323 & 0.279 \\ \hline
    Bi. CFS-AO &5 & \textbf{0.410} & 0.232 \\ \hline
    Bi. RF-IG&5&0.287& \textbf{0.430} \\ \hline
  \end{tabular}
}
\end{table}

\subsection{Random Forest}
\label{subsec:rf_res}

Once we determined our finalized hyperparameters, we ran the models and generated the performance results shown in \cref{tab:rf_perf_res}. The performance results are calculated as discussed in \cref{subsec:assessment} and the time results compare the build time, describing the time taken to actually build the model. In terms of categorical results, CFS-BA and RF-IG both outperform the full feature set, with RF-IG being strongest in the most metrics and CFS-BA being significantly faster. For binary classification, all models not using the full set are worse than building a model for categorical classification and using it solely for binary decision making.

\begin{table}[H]
\footnotesize
\caption{Random Forest Performance Results}\label{tab:rf_perf_res}
\centering
\resizebox{\columnwidth}{!}{
  \begin{tabular}{|l|c|c|c|c|c|} \hline
    Methodology&Time (s) & Accuracy & Precision & FAR & F1 \\ \hline
    Cat. Full & 6954 & 98.41\% & \textbf{90.45}\% & 22.78\% & 79.64\% \\ \hline
    Cat. CFS-BA & 5727 & \textbf{98.44}\%	& 89.5\%	& 21.26\%	& 81.02\% \\ \hline
    Cat. CFS-AO & \textbf{4146} & 98.40\%	& 88.60\%	& 21.62\%	& 80.21\% \\ \hline
    Cat. RF-IG & 10359 & \textbf{98.44}\%	& 89.3\%	& \textbf{19.79}\%	& \textbf{82.09}\% \\ \hline \hline
    Bi. Full & 5392 & \textbf{98.96}\% & \textbf{99.16}\% & \textbf{2.86}\%	& \textbf{98.12}\% \\ \hline
    Bi. CFS-BA & 2161 & 96.01\% & 91.49\% & 4.68\%	& 93.26\% \\ \hline
    Bi. CFS-AO & \textbf{1792} & 95.64\% & 90.69\% & 4.90\%	& 92.69\% \\ \hline
    Bi. RF-IG & 5493 & 97.19\% & 98.14\%	& 8.08\%	& 94.69\% \\ \hline
  \end{tabular}
}
\end{table}

\subsection{Deep Neural Network}
\label{subsec:dnn_res}

We run our DNN based upon the methodology declared in \cref{subsubsec:meth_clas_dnn}. Results for our performance are shown in \cref{tab:dnn_50_perf_res}. With regards to performance, we see that CFS-BA outperforms the full feature set in every binary metric. In categorical classification, CFS-BA gets the highest accuracy and the lowest FAR but is beat by the full set in F1 and the RF-IG model in precision.

\begin{table}[H]
\footnotesize
\caption{DNN Performance Results}\label{tab:dnn_50_perf_res}
\centering
\resizebox{\columnwidth}{!}{
  \begin{tabular}{|l|c|c|c|c|c|} \hline
    Methodology&Time (s) & Accuracy & Precision & FAR & F1 \\ \hline
    Cat. Full & \textbf{7689} & 98.01\% &79.38\% &29.66\% &70.67\% \\ \hline
    Cat. CFS-BA & 8413 & 98.29\% &84.52\% &29.44\% &71.63\% \\ \hline
    Cat. CFS-AO & 8125 & 98.11\% &81.03\% &30.08\% &71.21\% \\ \hline
    Cat. RF-IG & 13376 & \textbf{98.35}\% &\textbf{85.41}\% &\textbf{24.83}\% &\textbf{76.92}\% \\ \hline \hline
    Bi. Full & 9032 & \textbf{98.95}\% &\textbf{99.30}\% &\textbf{3.01}\% &\textbf{98.10}\% \\ \hline
    Bi. CFS-BA & 6136 & 93.77\% &95.80\% &17.89\% &87.13\% \\ \hline
    Bi. CFS-AO & \textbf{6103} & 93.65\% &95.32\% &17.99\% &86.92\% \\ \hline
    Bi. RF-IG & 9382 & 97.06\% &98.04\% &8.43\% &94.43\% \\ \hline
  \end{tabular}
}

\end{table}

Looking at the confusion matrices of RF and DNN, we see some patterns that may explain discrepancies. In both model types, DoS and BruteForce attacks were commonly mislabeled as one another. Web attacks, being the smallest and most difficult to catch, were actually more easily recognized than Infiltration attacks in RF models. DNN models nearly entirely missed Web attacks, and struggled just as much with Infiltration attacks. All other attacks performed relatively similarly for both model types. RF models were more inclined to mislabel to fewer types of labels, meaning confusion matrices with less 0's than their DNN alternatives.

\section{Discussion}
\label{sec:discussion}

Our models are able to achieve a maximum accuracy of 98.44\% on the CSE-CIC-IDS2018 dataset, using RF-IG with 14 of the 70 features for categorical classification. Both RF and DNN values show CFS-BA and RF-IG to outperform the full feature set. CFS-AO outperforms the full set for DNN, but not for RF. Using F1-score as a holistic metric, RF-IG raises 2.45\% and 6.25\% for RF and DNN respectively, while CFS-BA raises 1.38\% and 0.96\%. Using accuracy to determine applicability, RF-IG raises 0.34\% and 0.03\%, and CFS-BA raises 0.28\% and 0.03\%. 

The information from this paper suggests subset size or IG score may be more important than CFS. Contrasting CFS-BA to CFS-AO, the two benefits CFS-BA may have are a larger subset size and IG score. Of these two, IG score seems to be more likely to directly correlate to performance. The full feature set has the largest number of features and yet underperforms. CFS-BA has more features than RF-IG and yet generates comparable results for categorical classification and poorer results for binary classification. CFS-BA has an IG score 92.9\% of RF-IG's for categorical classification and performs much better than CFS-AO, having an IG score 51.6\% of RF-IG's.

Despite all of this, CFS-BA maintains advantages to RF-IG in the trade off between build time and performance. It builds in 62.9\% of the time for DNN and 55.3\% of the time for RF. Towards explainability, CFS-BA seems to be a strong method for funneling down a set of useful features. While not as direct as RF-IG, the models share many features, and so it can be used as a rough RF-IG estimate.

In the realm of explainability, RF-IG provides individual scores of importance while CFS only scores subsets. This means that RF-IG is a superior and more accurate method when making recommendations for a particular vulnerability of a site, while CFS may be more applicable towards reinforcing aspects of a site, such as showing where the heaviest loads may be distributed if an attack is slow to be caught. However, as shown in our time results, if someone wishes to update individualized features in a rapid manner, CFS-BA can be a relatively comparable alternative to RF-IG.

Fundamentally, CFS using Spearman's correlation tests monotonicity, which may be why it does not directly indicate performance. We know our data is highly skewed, and it may have too many non-monotonic relationships. This could be because factors have monotonic relationships with one another within a certain classification, but the blending of all points generates noise that distorts calculations. A rationale to this would be the performance of binary models being weaker than categorical models. To overcome this, future models could look at binary $M_s$ for each individual class, and take the union of all feature subsets. Another piece of data reinforcing this theory is that the 100-100-100 model was worse than the 50-25 model. This suggests either that the data is t0o sparse, or that the data appears sparse because the features selected are not an accurate reflection of the most beneficial features.

\section{Conclusion}
\label{sec:conclusion}

In short, this paper added results for DNN and RF performance for CSE-CIC-IDS2018 using the full feature set, CFS-BA, CFS-AO, and RF-IG for the same model hyperparameters. The use of CFS-BA and RF-IG indisputably benefited the model. It removed 63\%+ of features and significant improved model performance.

There are many directions the work can go in to increase its applicability in the field. CNN structure could be integrated as a replacement for DNN to attempt to account for the highly skewed distribution in the data as well as clear the feature obfuscation present. Shapley game theory scores could also be used to further improve the explainability of the models and provide more insight into why RF-IG excels where CFS may not. Lastly, scores other than CFS could be applied for filter comparison, or methods of correlation such as Pearson's or individualized Spearman's may be a better application of CFS.

\bibliographystyle{IEEEtran}
\bibliography{IEEEabrv,references}

\begin{thebibliography}{10}
\providecommand{\url}[1]{#1}
\csname url@samestyle\endcsname
\providecommand{\newblock}{\relax}
\providecommand{\bibinfo}[2]{#2}
\providecommand{\BIBentrySTDinterwordspacing}{\spaceskip=0pt\relax}
\providecommand{\BIBentryALTinterwordstretchfactor}{4}
\providecommand{\BIBentryALTinterwordspacing}{\spaceskip=\fontdimen2\font plus
\BIBentryALTinterwordstretchfactor\fontdimen3\font minus
  \fontdimen4\font\relax}
\providecommand{\BIBforeignlanguage}[2]{{%
\expandafter\ifx\csname l@#1\endcsname\relax
\typeout{** WARNING: IEEEtran.bst: No hyphenation pattern has been}%
\typeout{** loaded for the language `#1'. Using the pattern for}%
\typeout{** the default language instead.}%
\else
\language=\csname l@#1\endcsname
\fi
#2}}
\providecommand{\BIBdecl}{\relax}
\BIBdecl

\bibitem{noauthor_malware_nodate}
\BIBentryALTinterwordspacing
Malware statistics \& trends report {\textbar} {AV}-{TEST}. [Online].
  Available: \url{https://www.av-test.org/en/statistics/malware/}
\BIBentrySTDinterwordspacing

\bibitem{noauthor_cost_2021}
\BIBentryALTinterwordspacing
Cost of a data breach report 2021. [Online]. Available:
  \url{https://www.ibm.com/security/data-breach}
\BIBentrySTDinterwordspacing

\bibitem{xin_machine_2018}
Y.~Xin \emph{et~al.}, ``Machine learning and deep learning methods for
  cybersecurity,'' vol.~6, pp. 35\,365--35\,381, conference Name: {IEEE}
  Access.

\bibitem{sarker_cybersecurity_2020}
\BIBentryALTinterwordspacing
I.~H. Sarker, A.~S.~M. Kayes, S.~Badsha, H.~Alqahtani, P.~Watters, and A.~Ng,
  ``Cybersecurity data science: an overview from machine learning
  perspective,'' vol.~7, no.~1, p.~41. [Online]. Available:
  \url{https://doi.org/10.1186/s40537-020-00318-5}
\BIBentrySTDinterwordspacing

\bibitem{farhan_performance_2020}
\BIBentryALTinterwordspacing
R.~I. Farhan, A.~T. Maolood, and N.~F. Hassan, ``Performance analysis of
  flow-based attacks detection on {CSE}-{CIC}-{IDS}2018 dataset using deep
  learning,'' vol.~20, no.~3, pp. 1413--1418, number: 3. [Online]. Available:
  \url{https://ijeecs.iaescore.com/index.php/IJEECS/article/view/22288}
\BIBentrySTDinterwordspacing

\bibitem{zhou_building_2020}
\BIBentryALTinterwordspacing
Y.~Zhou, G.~Cheng, S.~Jiang, and M.~Dai, ``Building an efficient intrusion
  detection system based on feature selection and ensemble classifier,'' vol.
  174, p. 107247. [Online]. Available: \url{http://arxiv.org/abs/1904.01352}
\BIBentrySTDinterwordspacing

\bibitem{james_introduction_2021}
G.~James, D.~Witten, T.~Hastie, and R.~Tibshirani, \emph{An Introduction to
  Statistical Learning with Applications in R}, 2nd~ed.

\bibitem{leevy_survey_2020}
\BIBentryALTinterwordspacing
J.~L. Leevy and T.~M. Khoshgoftaar, ``A survey and analysis of intrusion
  detection models based on {CSE}-{CIC}-{IDS}2018 big data,'' vol.~7, no.~1, p.
  104. [Online]. Available: \url{https://doi.org/10.1186/s40537-020-00382-x}
\BIBentrySTDinterwordspacing

\bibitem{mahdavifar_application_2019}
\BIBentryALTinterwordspacing
S.~Mahdavifar and A.~A. Ghorbani, ``Application of deep learning to
  cybersecurity: A survey,'' vol. 347, pp. 149--176. [Online]. Available:
  \url{https://www.sciencedirect.com/science/article/pii/S0925231219302954}
\BIBentrySTDinterwordspacing

\bibitem{akashdeep_feature_2017}
\BIBentryALTinterwordspacing
{Akashdeep}, I.~Manzoor, and N.~Kumar, ``A feature reduced intrusion detection
  system using {ANN} classifier,'' vol.~88, pp. 249--257. [Online]. Available:
  \url{https://www.sciencedirect.com/science/article/pii/S0957417417304748}
\BIBentrySTDinterwordspacing

\bibitem{li_building_2020}
\BIBentryALTinterwordspacing
X.~Li, W.~Chen, Q.~Zhang, and L.~Wu, ``Building auto-encoder intrusion
  detection system based on random forest feature selection,'' vol.~95, p.
  101851. [Online]. Available:
  \url{https://www.sciencedirect.com/science/article/pii/S0167404820301231}
\BIBentrySTDinterwordspacing

\bibitem{yang_bat_2012}
X.-S. Yang and A.~Gandomi, ``Bat algorithm: A novel approach for global
  engineering optimization,'' vol.~29.

\bibitem{yang_bat_2013}
\BIBentryALTinterwordspacing
X.-S. Yang and X.~He, ``Bat algorithm: literature review and applications,''
  vol.~5, no.~3, pp. 141--149. [Online]. Available:
  \url{https://doi.org/10.1504/IJBIC.2013.055093}
\BIBentrySTDinterwordspacing

\bibitem{fatani_advanced_2022}
\BIBentryALTinterwordspacing
A.~Fatani, A.~Dahou, M.~A.~A. Al-qaness, S.~Lu, and M.~Abd~Elaziz, ``Advanced
  feature extraction and selection approach using deep learning and aquila
  optimizer for {IoT} intrusion detection system,'' vol.~22, no.~1, p. 140,
  number: 1 Publisher: Multidisciplinary Digital Publishing Institute.
  [Online]. Available: \url{https://www.mdpi.com/1424-8220/22/1/140}
\BIBentrySTDinterwordspacing

\bibitem{blakely_evaluation_2018}
L.~Blakely, M.~Reno, and R.~Broderick, ``Evaluation and comparison of machine
  learning techniques for rapid qsts simulations,'' 07 2018.

\bibitem{noauthor_deep-learning_nodate}
\BIBentryALTinterwordspacing
Deep-learning: investigating deep neural networks hyper-parameters and
  comparison of performance to shallow methods for modeling bioactivity data
  {\textbar} journal of cheminformatics {\textbar} full text. [Online].
  Available:
  \url{https://jcheminf.biomedcentral.com/articles/10.1186/s13321-017-0226-y}
\BIBentrySTDinterwordspacing

\bibitem{hall1999correlation}
M.~A. Hall \emph{et~al.}, ``Correlation-based feature selection for machine
  learning,'' 1999.

\bibitem{ahmad_network_2021}
\BIBentryALTinterwordspacing
Z.~Ahmad, A.~Shahid~Khan, C.~Wai~Shiang, J.~Abdullah, and F.~Ahmad, ``Network
  intrusion detection system: A systematic study of machine learning and deep
  learning approaches,'' vol.~32, no.~1, p. e4150, \_eprint:
  https://onlinelibrary.wiley.com/doi/pdf/10.1002/ett.4150. [Online].
  Available: \url{https://onlinelibrary.wiley.com/doi/abs/10.1002/ett.4150}
\BIBentrySTDinterwordspacing

\bibitem{noauthor_ids_nodate}
\BIBentryALTinterwordspacing
{IDS} 2018 {\textbar} datasets {\textbar} research {\textbar} canadian
  institute for cybersecurity {\textbar} {UNB}. [Online]. Available:
  \url{https://www.unb.ca/cic/datasets/ids-2018.html}
\BIBentrySTDinterwordspacing

\bibitem{noauthor_realistic_nodate}
\BIBentryALTinterwordspacing
A realistic cyber defense dataset ({CSE}-{CIC}-{IDS}2018) - registry of open
  data on {AWS}. [Online]. Available:
  \url{https://registry.opendata.aws/cse-cic-ids2018/}
\BIBentrySTDinterwordspacing

\bibitem{kaggle_ids_nodate}
\BIBentryALTinterwordspacing
{IDS} 2018 intrusion {CSVs} ({CSE}-{CIC}-{IDS}2018). [Online]. Available:
  \url{https://www.kaggle.com/datasets/solarmainframe/ids-intrusion-csv}
\BIBentrySTDinterwordspacing

\bibitem{scikit-learn}
F.~Pedregosa \emph{et~al.}, ``Scikit-learn: Machine learning in {P}ython,''
  \emph{Journal of Machine Learning Research}, vol.~12, pp. 2825--2830, 2011.

\bibitem{thieu_nguyen_2020_3711949}
\BIBentryALTinterwordspacing
N.~V. Thieu, ``A collection of the state-of-the-art meta-heuristics algorithms
  in python: Mealpy,'' 2020. [Online]. Available:
  \url{https://doi.org/10.5281/zenodo.3711948}
\BIBentrySTDinterwordspacing

\bibitem{tensorflow2015-whitepaper}
\BIBentryALTinterwordspacing
M.~Abadi \emph{et~al.}, ``{TensorFlow}: Large-scale machine learning on
  heterogeneous systems,'' 2015, software available from tensorflow.org.
  [Online]. Available: \url{https://www.tensorflow.org/}
\BIBentrySTDinterwordspacing

\end{thebibliography}

\end{document}